\documentclass[pdflatex,sn-basic]{sn-jnl}

\usepackage{graphicx}%
\usepackage{multirow}%
\usepackage{amsmath,amssymb,amsfonts}%
\usepackage{amsthm}%
\usepackage{mathrsfs}%
\usepackage[title]{appendix}%
\usepackage{xcolor}%
\usepackage{textcomp}%
\usepackage{manyfoot}%
\usepackage{booktabs}%
\usepackage{algorithm}%
\usepackage{algorithmicx}%
\usepackage{algpseudocode}%
\usepackage{listings}%
\usepackage{subcaption}

\usepackage{tikz}
\usepackage{xcolor}
\usetikzlibrary{positioning, arrows.meta}

\usepackage{booktabs}
\usepackage{colortbl}
\usepackage{xcolor}
\usepackage{array}
\usepackage{multirow}
\definecolor{anchorColor}{RGB}{29,158,117}
\definecolor{pathColor}{RGB}{55,138,221}
\definecolor{neighColor}{RGB}{230,180,20}
\colorlet{newColColor}{neighColor!35}
\colorlet{newColBorder}{neighColor!80!black}
\colorlet{selColColor}{pathColor!22}
\colorlet{selColBorder}{pathColor!80!black}

\theoremstyle{thmstyleone}%
\theoremstyle{thmstyletwo}%

\theoremstyle{thmstylethree}%

\begin{document}


\title[\texttt{path_boost}]{\texttt{path\_boost}: A Python Package for Interpretable Graph-Level Prediction using Path-Based Gradient Boosting}


\author*[1,]{\fnm{Claudio} \sur{Meggio}}\email{claudm@math.uio.no}

\author[1]{\fnm{Johan} \sur{Pensar}}\email{johanpen@math.uio.no}

\author[1]{\fnm{Riccardo} \sur{De Bin}}\email{debin@math.uio.no}

\affil[1]{\orgdiv{Department of Mathematics}, \orgname{University of Oslo}, \orgaddress{ \city{Oslo}, \postcode{0371}, \state{Oslo}, \country{Norway}}}


\abstract{
  We present \texttt{path\_boost}, a Python package for interpretable supervised learning on graph-structured input data. The package implements PathBoost, a gradient boosting algorithm that automatically discovers predictive labeled paths within graphs during the learning process. Unlike graph neural networks, which are generally difficult to interpret, PathBoost produces an additive prediction model over path-based features that explicitly reveals which substructures drive predictions. To avoid an exhaustive enumeration of all possible paths, the algorithm iteratively selects and extends paths during learning based on their predictive power, using boosting to combine weak learners into a strong ensemble. The package supports both regression and binary classification. Key features include compatibility with \texttt{scikit-learn} workflows, support for custom base learners and selectors, automatic starting node selection, parallel training across anchor nodes, and built-in variable importance computation. We demonstrate PathBoost on molecular property prediction of transition metal compounds, where atoms serve as nodes and bonds as edges, and further benchmark PathBoost against an established graph neural network and a graph kernel method across six molecular datasets. The package is available on PyPI and GitHub under an open-source license.
}


\maketitle

\section[Introduction]{Introduction} \label{sec:intro}

Graph-structured data are increasingly prevalent across scientific and industrial domains, including molecular chemistry, materials science, and social network analysis. Unlike traditional tabular data, graphs capture relationships and interactions between entities, making them a powerful yet challenging data structure for machine learning. In particular, learning predictive models from graphs requires methods that can extract meaningful features from their structure and, ideally, do so in an interpretable way.

Among the existing approaches for this prediction task, the most common are Graph Neural Network (GNN) and graph kernel methods. GNNs have become the dominant paradigm for learning from graph data, achieving strong empirical results on many benchmarks \citep{gilmer2017neural, hu2020ogb}. However, GNNs present several practical limitations: they are generally considered difficult to interpret \citep{Wu2021} and often require a lot of training data to generalize effectively \citep{Ding2024}. Classical graph kernel methods paired with support vector machines remain competitive on small-scale datasets \citep{Kriege2020}, but rely on fixed feature representations that do not adapt during learning, and the resulting models offer limited interpretability in practice.

Previous work by \citet{meggio2024pathboost} introduced a path-based boosting framework for regression on molecular graphs of transition metal compounds, where a designated metal-centre node, called the anchor node, is used as the starting node for path exploration. \citet{meggio_2nd} extended this framework in several directions: incorporating multiple node and edge attributes into the path feature space via a two-step base learner that separates path selection from path fitting; introducing automatic anchor node selection based on categorical attribute diversity; and adapting the algorithm to binary classification through the logistic loss. These works also formalized absolute and relative variants of path-level variable importance, and demonstrated competitive performance against graph neural networks and kernel methods on twelve TUDataset benchmarks.

In this paper, we present \texttt{path\_boost}, an open source Python package that implements the PathBoost algorithm developed in \citet{meggio2024pathboost} and \citet{meggio_2nd}. In addition, \texttt{path\_boost} introduces two methodological extensions: a multi-anchor parallel architecture that trains independent sub-models for each anchor label and aggregates their predictions, and an enhanced variable importance framework that includes a correlation adjustment to account for structural dependencies between nested paths. The implementation follows \texttt{scikit-learn} conventions, implementing \texttt{fit()}, \texttt{predict()}, \texttt{get\_params()}, and \texttt{set\_params()} methods, enabling compatibility with standard tools such as \texttt{GridSearchCV} for hyperparameter tuning. Graphs are represented using the \texttt{networkx} library \citep{hagberg2008exploring}.

The remainder of this paper is organized as follows. Section~\ref{sec:preliminaries} introduces key terminology and defines the problem setting. Section~\ref{sec:algorithm} describes the algorithm, including the boosting procedure with a step-by-step illustration, and variable importance measures. Section~\ref{sec:implementation} covers software design and implementation details. Section~\ref{sec:application} provides a tutorial demonstrating the package on a molecular property prediction task using the tmQMg dataset. Section~\ref{sec:graphbench} reports a benchmark comparison against GINE and WL~+~SVR across six molecular datasets. Section~\ref{sec:summary} concludes with some remarks on the performance of the package.

\section{Preliminaries}\label{sec:preliminaries}
\subsection{Terminology}\label{sec:terminology}

We introduce the key terms used throughout this paper:
\begin{itemize}
    \item \emph{Labeled path}: an ordered sequence of anchor-attribute values $(l_1, l_2, \ldots, l_m)$ corresponding to a walk through the graph where consecutive nodes are connected by edges.
    \item \emph{Boosting Matrix (BM)}: a frequency matrix where each column corresponds to a labeled path and each row to a graph. Entry $(i, u)$ records how many times labeled path $u$ occurs in graph $G_i$.
    \item \emph{Extended Boosting Matrix (EBM)}: an extended version of the BM that includes, for each labeled path, the averaged numerical node and edge attributes along that path and all its prefixes.
    \item \emph{Prefix}: a labeled path $q = (q_1, \ldots, q_k)$ is a prefix of path $p = (p_1, \ldots, p_m)$ if $k < m$ and $q_i = p_i$ for all $i \in \{1, \ldots, k\}$; that is, $q$ is an initial sub-sequence of $p$.
    \item \emph{Selector model}: a simple model (by default, a decision tree stump) fitted on the BM to identify the most informative labeled path at each boosting iteration.
    \item \emph{Base learner}: the model fitted on the EBM columns relative to the selected path to produce the boosting update. By default, a decision tree regressor.
    \item \emph{Anchor attribute} (\texttt{anchor\_nodes\_label\_name}): the categorical node attribute that determines where paths can start. Each of its distinct values picks out the nodes that carry that value, and paths are allowed to begin from such nodes. For example, in molecular graphs the anchor attribute might be \texttt{"feature\_atomic\_number"}: this means we use the atomic number to distinguish the different anchor nodes.
    \item \emph{Anchor label} (\texttt{list\_anchor\_nodes\_labels}): a specific value of the anchor attribute that identifies a class of nodes as starting points for path exploration. Only nodes whose anchor attribute equals a selected anchor label are used as path origins. For example, in a molecular graph where the anchor attribute is the atomic number, anchor labels \texttt{[25, 47, 78]} restrict path exploration to start from manganese, silver, and platinum atoms, respectively.
\end{itemize}

\subsection{Problem setting}\label{sec:problem}

We are given a dataset of $N$ graphs $\mathcal{G} = \{G_1, G_2, \ldots, G_N\}$, where each graph $G_i = (V_i, E_i, A_{V_i}, A_{E_i})$ consists of a set of nodes $V_i$, a set of edges $E_i \subset V_i \times V_i$, a node attribute function $A_{V_i} \colon V_i \to \mathbb{R}^{q_V}$ mapping each node to a vector of $q_V$ attributes, and an edge attribute function $A_{E_i} \colon E_i \to \mathbb{R}^{q_E}$ mapping each edge to a vector of $q_E$ attributes. Moreover, each graph is associated with a target value $y_i\in\mathcal{Y}$. The goal is to learn a function $f \colon \mathcal{G} \to \mathcal{Y}$ that predicts $y_i$ from $G_i$. The package supports regression, where $\mathcal{Y} = \mathbb{R}$, and binary classification, where $\mathcal{Y} = \{0, 1\}$.


A key requirement is that at least one node attribute is \emph{categorical}, which means that it takes finitely many distinct values. One such attribute is designated as the \emph{anchor attribute}: its distinct values, or a chosen subset of them, form the set of \emph{anchor labels}, the values from which path exploration is allowed to start. A labeled path is then a sequence of values of this anchor attribute. Apart from this role, the anchor attribute is treated like any other --- every attribute contributes its averaged value to the EBM (Section~\ref{subsec:ebm_structure}). For example, in molecular graphs where nodes represent atoms and edges represent bonds, the chemical element is the anchor attribute, and each element of the periodic table (e.g., carbon, nitrogen, oxygen) is an anchor label. In the implementation, elements are identified by their atomic number, so these categorical labels appear as integers (e.g., 6, 7, 8 for carbon, nitrogen, oxygen).

The method is particularly suited to problems where the target depends on local structural patterns within the graphs; however, additional numerical node and edge attributes (e.g., atomic mass, bond strength) can be incorporated into the feature space to enrich the path-based representation, as described in Section~\ref{sec:algorithm}.

\section{Algorithm}\label{sec:algorithm}

PathBoost constructs a feature space from graph-structured data that can be used by standard supervised learning algorithms. It does so by iteratively discovering the most informative labeled paths within graphs via gradient boosting.

\subsection{The boosting procedure}\label{subsec:boosting}

The core algorithm, summarized in Algorithm~\ref{alg:spb}, proceeds as follows to minimize a given loss function, $\ell$. Given training graphs and associated target values, the boosting matrix (BM) is initialized with single-node paths corresponding to the anchor labels. At each iteration: (i) the selector identifies the most predictive path from the BM, (ii) the extended boosting matrix (EBM) is constructed for that path along with its prefixes (if additional attributes are available), (iii) a base learner is fitted to the pseudo-residuals using the EBM features, and (iv) the model predictions are updated. When a path is selected for the first time, the BM is expanded with all one-step extensions of that path found in the training data. This lazy expansion avoids the combinatorial cost of enumerating all possible paths upfront.

\begin{algorithm}[htbp]
\caption{SequentialPathBoost Fitting Procedure}
\label{alg:spb}
\begin{algorithmic}[1]
\Require Training graphs $\mathcal{G}$, targets $\mathbf{y}$, set of anchor labels $\mathcal{A}$, anchor attribute $L$, number of iterations $T$, learning rate $\eta$
\State \textbf{Initialization:}
    \State \hspace{1em} Initialize the BM with features from anchor nodes in $\mathcal{G}$ using $\mathcal{A}$ and $L$
    \State \hspace{1em} Set initial prediction $\hat{\mathbf{y}}^{(0)} \gets F_0(\mathbf{y})$, where $F_0 = \bar{y} = \frac{1}{N}\sum_{i=1}^N y_i$ for regression and $F_0 = \log\!\left(\bar{y}/(1-\bar{y})\right)$ for classification
\For{$t = 1$ to $T$}
    \State Compute pseudo-residuals: $\mathbf{r}^{(t)} \gets -\nabla_{\hat{\mathbf{y}}}\,\ell(\mathbf{y},\, \hat{\mathbf{y}}^{(t-1)})$
    \State \textit{Path selection:} Use the selector model on the BM to identify the most informative labeled path $p^*$ with respect to $\mathbf{r}^{(t)}$
    \State \textit{Feature construction:} Build the EBM columns for $p^*$ and all its prefixes
    \State \textit{Fit base learner:} Train base learner $h^{(t)}$ on the EBM columns of $p^*$, targeting $\mathbf{r}^{(t)}$
    \State \textit{Update predictions:} $\hat{\mathbf{y}}^{(t)} \gets \hat{\mathbf{y}}^{(t-1)} + \eta \cdot h^{(t)}(\text{EBM}_{p^*})$
    \If{$p^*$ has not been selected before}
        \State \textit{Expand BM:} Add columns for all one-step extensions of $p^*$ found in $\mathcal{G}$
    \EndIf
\EndFor
\end{algorithmic}
\end{algorithm}

A key design choice is the separation between path \emph{selection} and path \emph{fitting}. The selector operates only on the BM, while the base learner operates on the full EBM. This separation is motivated by two primary considerations. First, \emph{computational efficiency}: the BM is a compact matrix of integer counts (one column per candidate path, one row per graph) and can be evaluated very quickly by a simple model such as a decision stump. The EBM, by contrast, includes all averaged node and edge attributes for each path and its prefixes, making it much larger. Using the base learner directly for path selection would require fitting this richer model once per candidate path at each iteration, which would be prohibitively expensive. Second, \emph{flexibility}: because path selection depends only on structural frequency counts and not on attribute values, the selector and the base learner can be chosen independently. The selector can stay a fast, simple model, while the base learner that fits the attribute-rich EBM can be any compatible estimator, from a shallow tree to a linear or gradient-boosted model, without changing how paths are selected. This modularity also preserves interpretability: the selected path provides a clear structural interpretation regardless of how the base learner uses the associated attributes.

In practice, the number of iterations, $T$, is set through the \texttt{n\_iter} parameter. This quantity is the most important hyperparameter: too few iterations and the model underfits, too many and it overfits. The package supports two possible strategies for selecting it. The first is to treat \texttt{n\_iter} as any other hyperparameter and tune it via cross-validation. The second is to fix a upper bound for \texttt{n\_iter} and let the algorithm decide when to stop. To do so, two criteria are available: \texttt{patience} stops training after a given number of consecutive iterations without improvement, and \texttt{target\_error} stops training once the evaluation MSE falls below a user-defined threshold. Section~\ref{subsec:cross_validation} discusses how to use them in practice.

\subsection{Illustration: boosting iteration on a molecular graph}\label{subsec:illustration}

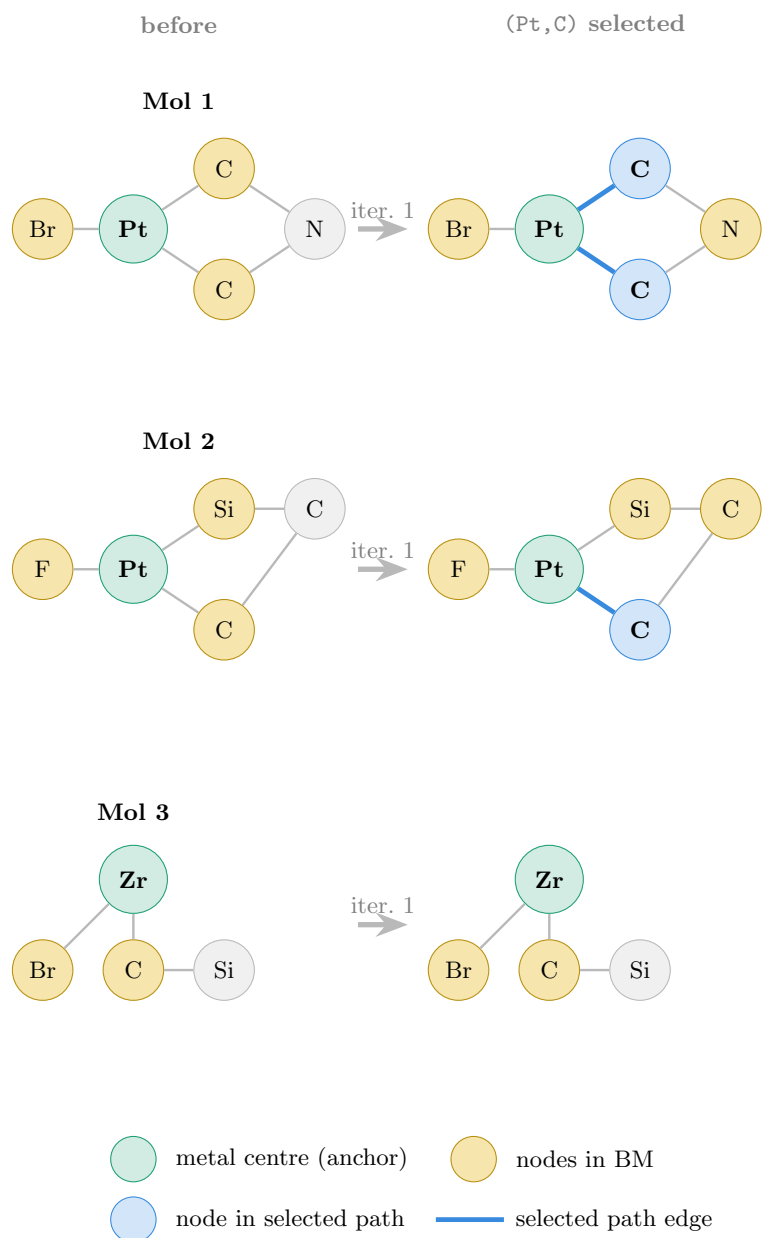
\begin{figure}[htbp]
\centering
\small
 
\begin{tikzpicture}[
    font=\small,
    >=Stealth,
    metal-atom/.style  ={circle, draw=anchorColor,         fill=anchorColor!20,
                         minimum size=9mm, inner sep=0pt, font=\bfseries\small},
    neigh-atom/.style  ={circle, draw=neighColor!80!black, fill=neighColor!35,
                         minimum size=8mm, inner sep=0pt, font=\small},
    plain-atom/.style  ={circle, draw=gray!55,             fill=gray!12,
                         minimum size=8mm, inner sep=0pt, font=\small},
    path-hl/.style     ={circle, draw=pathColor,           fill=pathColor!22,
                         minimum size=8mm, inner sep=0pt, font=\bfseries\small},
    plain-edge/.style  ={draw=gray!55,   line width=0.9pt},
    path-edge/.style   ={draw=pathColor, line width=1.8pt},
    mol-arrow/.style   ={->, line width=2pt, gray!55,
                         shorten >=2pt, shorten <=2pt},
]
 
\node[metal-atom] (Pt1L) at (0.0,  0.8) {Pt};
\node[neigh-atom] (C1La) at (1.2,  1.6) {C};
\node[plain-atom] (N1L)  at (2.4,  0.8) {N};
\node[neigh-atom] (C1Lb) at (1.2,  0.0) {C};
\node[neigh-atom] (Br1L) at (-1.2, 0.8) {Br};
 
\draw[plain-edge] (Pt1L) -- (C1La);
\draw[plain-edge] (Pt1L) -- (C1Lb);
\draw[plain-edge] (Pt1L) -- (Br1L);
\draw[plain-edge] (C1La) -- (N1L);
\draw[plain-edge] (C1Lb) -- (N1L);
 
\node[font=\bfseries] at (0.6, 2.5) {Mol 1};
\draw[mol-arrow] (2.9, 0.8) -- (3.7, 0.8);
\node[font=\scriptsize, gray, above=1pt] at (3.3, 0.8) {iter.\ 1};
 
\begin{scope}[xshift=5.5cm]
\node[metal-atom]  (Pt1R) at (0.0,  0.8) {Pt};
\node[path-hl]     (C1Ra) at (1.2,  1.6) {C};
\node[neigh-atom]  (N1R)  at (2.4,  0.8) {N};
\node[path-hl]     (C1Rb) at (1.2,  0.0) {C};
\node[neigh-atom]  (Br1R) at (-1.2, 0.8) {Br};
 
\draw[path-edge]  (Pt1R) -- (C1Ra);
\draw[path-edge]  (Pt1R) -- (C1Rb);
\draw[plain-edge] (Pt1R) -- (Br1R);
\draw[plain-edge] (C1Ra) -- (N1R);
\draw[plain-edge] (C1Rb) -- (N1R);
\end{scope}
 
\begin{scope}[yshift=-4.5cm]
\node[metal-atom] (Pt2L) at (0.0,  0.8) {Pt};
\node[neigh-atom] (Si2L) at (1.2,  1.6) {Si};
\node[plain-atom] (C2La) at (2.4,  1.6) {C};
\node[neigh-atom] (C2Lb) at (1.2,  0.0) {C};
\node[neigh-atom] (F2L)  at (-1.2, 0.8) {F};
 
\draw[plain-edge] (Pt2L) -- (Si2L);
\draw[plain-edge] (Pt2L) -- (F2L);
\draw[plain-edge] (Pt2L) -- (C2Lb);
\draw[plain-edge] (Si2L) -- (C2La);
\draw[plain-edge] (C2Lb) -- (C2La);
 
\node[font=\bfseries] at (0.6, 2.5) {Mol 2};
\draw[mol-arrow] (2.9, 0.8) -- (3.7, 0.8);
\node[font=\scriptsize, gray, above=1pt] at (3.3, 0.8) {iter.\ 1};
 
\begin{scope}[xshift=5.5cm]
\node[metal-atom]  (Pt2R) at (0.0,  0.8) {Pt};
\node[neigh-atom]  (Si2R) at (1.2,  1.6) {Si};
\node[neigh-atom]  (C2Ra) at (2.4,  1.6) {C};
\node[path-hl]     (C2Rb) at (1.2,  0.0) {C};
\node[neigh-atom]  (F2R)  at (-1.2, 0.8) {F};
 
\draw[plain-edge] (Pt2R) -- (Si2R);
\draw[plain-edge] (Pt2R) -- (F2R);
\draw[path-edge]  (Pt2R) -- (C2Rb);
\draw[plain-edge] (Si2R) -- (C2Ra);
\draw[plain-edge] (C2Rb) -- (C2Ra);
\end{scope}
\end{scope}
 
\begin{scope}[yshift=-9.0cm]
\node[metal-atom] (Zr3L) at (0.0,  1.2) {Zr};
\node[neigh-atom] (Br3L) at (-1.2, 0.0) {Br};
\node[neigh-atom] (C3L)  at (0.0,  0.0) {C};
\node[plain-atom] (Si3L) at (1.2,  0.0) {Si};
 
\draw[plain-edge] (Zr3L) -- (Br3L);
\draw[plain-edge] (Zr3L) -- (C3L);
\draw[plain-edge] (C3L)  -- (Si3L);
 
\node[font=\bfseries] at (0.0, 2.1) {Mol 3};
\draw[mol-arrow] (2.9, 0.6) -- (3.7, 0.6);
\node[font=\scriptsize, gray, above=1pt] at (3.3, 0.6) {iter.\ 1};
 
\begin{scope}[xshift=5.5cm]
\node[metal-atom] (Zr3R) at (0.0,  1.2) {Zr};
\node[neigh-atom] (Br3R) at (-1.2, 0.0) {Br};
\node[neigh-atom] (C3R)  at (0.0,  0.0) {C};
\node[plain-atom] (Si3R) at (1.2,  0.0) {Si};
 
\draw[plain-edge] (Zr3R) -- (Br3R);
\draw[plain-edge] (Zr3R) -- (C3R);
\draw[plain-edge] (C3R)  -- (Si3R);
\end{scope}
\end{scope}
 
\node[font=\bfseries\small, gray] at (0.6, 3.5) {before};
\node[font=\bfseries\small, gray] at (6.1, 3.5) {\texttt{(Pt,C)} selected};
 
\begin{scope}[yshift=-11.5cm]
\node[metal-atom,  minimum size=6mm] (lM) at (0.0,  0.0) {};
\node[right=4pt of lM, font=\scriptsize] {metal centre (anchor)};
 
\node[neigh-atom,  minimum size=6mm] (lN) at (4.5,  0.0) {};
\node[right=4pt of lN, font=\scriptsize] {nodes in BM};
 
\node[path-hl,     minimum size=6mm] (lP) at (0.0, -0.8) {};
\node[right=4pt of lP, font=\scriptsize] {node in selected path};
 
 
\draw[path-edge]  (4.0, -0.8) -- (4.9, -0.8);
\node[right=4pt, font=\scriptsize] at (4.8,  -0.8) {selected path edge};
 
\end{scope}
 
\end{tikzpicture}
 
\caption{%
  Illustration of path selection on three toy molecules. \textbf{Left column:} input molecular graphs with \textcolor{anchorColor}{\textbf{teal}} metal-centre anchor nodes and \textcolor{neighColor!80!black}{\textbf{yellow}} nodes corresponding to the initial Boosting Matrix columns (direct neighbours of the anchor). \textbf{Right column:} after the selector identifies \texttt{(Pt,\,C)} as the most informative path, the selected C nodes and Pt--C edges are highlighted in \textcolor{pathColor}{\textbf{blue}}; \textcolor{neighColor!80!black}{\textbf{yellow}} nodes are all nodes appearing as terminal nodes of BM columns at this iteration.  Molecule~3 contains no Pt anchor, so no path is highlighted.
}
\label{fig:path_selection}
\end{figure}

\begin{figure}[htbp]
\centering
\small
 
{\bfseries Boosting Matrix (BM)}\\[4pt]
\renewcommand{\arraystretch}{1.35}
\setlength{\tabcolsep}{4pt}

\begin{tabular}{l | c c c c c c | >{\columncolor{newColColor}}c
                                  >{\columncolor{newColColor}}c}
  \toprule
  & \multicolumn{6}{c|}{\textit{existing}}
  & \multicolumn{2}{c}{\textcolor{newColBorder}{\textit{ new, extensions of \texttt{(Pt,C)}}}} \\
  \cmidrule(lr){2-7} \cmidrule(l){8-9}
  & \texttt{(Pt)} & \texttt{(Zr)} & \texttt{(Pt,C)} & \texttt{(Pt,Br)} & \texttt{(Pt,Si)} & \texttt{(Zr,Br)}
  & \texttt{\textcolor{newColBorder}{(Pt,C,N)}}
  & \texttt{\textcolor{newColBorder}{(Pt,C,C)}} \\
  \midrule
  Mol 1 & 1 & 0 & 2 & 1 & 0 & 0
    & \textcolor{newColBorder}{2}
    & \textcolor{newColBorder}{0} \\
  Mol 2 & 1 & 0 & 1 & 0 & 1 & 0
    & \textcolor{newColBorder}{0}
    & \textcolor{newColBorder}{1} \\
  Mol 3 & 0 & 1 & 0 & 0 & 0 & 1
    & \textcolor{newColBorder}{0}
    & \textcolor{newColBorder}{0} \\
  \bottomrule
\end{tabular}

\bigskip
 
{\bfseries Selected Extended Boosting Matrix (EBM)}\\[4pt]
\renewcommand{\arraystretch}{1.35}
\setlength{\tabcolsep}{6pt}
\begin{tabular}{l | c c | >{\columncolor{selColColor}}c
                            >{\columncolor{selColColor}}c
                            >{\columncolor{selColColor}}c}
  \toprule
  & \multicolumn{2}{c|}{\textit{prefix: \texttt{(Pt)}}}
  & \multicolumn{3}{c}{\textcolor{selColBorder}{\textit{new --- selected path: \texttt{(Pt,C)}}}} \\
  \cmidrule(lr){2-3} \cmidrule(l){4-6}
  & \textbf{n\_times}
  & \textbf{atomic\_mass}
  & \textbf{\textcolor{selColBorder}{n\_times}}
  & \textbf{\textcolor{selColBorder}{atomic\_mass}}
  & \textbf{\textcolor{selColBorder}{bond\_order}} \\
  \midrule
  Mol 1 & 1 & 195.1
    & \textcolor{selColBorder}{2}
    & \textcolor{selColBorder}{12.0}
    & \textcolor{selColBorder}{1.0} \\
  Mol 2 & 1 & 195.1
    & \textcolor{selColBorder}{1}
    & \textcolor{selColBorder}{12.0}
    & \textcolor{selColBorder}{1.0} \\
  Mol 3 & 0 & ---
    & \textcolor{selColBorder}{0}
    & \textcolor{selColBorder}{---}
    & \textcolor{selColBorder}{---} \\
  \bottomrule
\end{tabular}
 
\caption{%
  Boosting Matrix (BM, top) and Selected Extended Boosting Matrix (EBM, bottom)
  at the second boosting iteration, after \texttt{(Pt,C)} has been selected.
  In the BM, existing columns include the anchor paths, the previously
  selected path~\texttt{(Pt,C)}, and path extensions of other anchors (e.g.\ \texttt{(Zr,$\cdot$)}) omitted for brevity~($\ldots$); \colorbox{newColColor}{\textcolor{newColBorder}{shaded columns}} are the
  two-step extensions of~\texttt{(Pt,C)} added at this iteration.
  In the EBM, the prefix group~\texttt{(Pt)} columns are pre-existing (unshaded),
  while the \colorbox{selColColor}{\textcolor{selColBorder}{shaded}}
  selected-path group~\texttt{(Pt,C)} is new: it records the number of occurrences,
  the averaged \texttt{atomic\_mass} of the last atom C, and the averaged
  \texttt{bond\_order} of the Pt--C edge.
  Dashes indicate paths absent in a graph.
}
\label{tab:bm_ebm}
\end{figure}

To illustrate how PathBoost discovers paths, we present a step-by-step walkthrough using the three toy molecules shown in Figure~\ref{fig:path_selection}, with platinum and zirconium (\texttt{(Pt)} and \texttt{(Zr)}) as the anchor labels.
In these molecular graphs, nodes correspond to atoms and edges to bonds; for readability, atoms are labelled by their atomic symbol rather than their atomic number. The algorithm proceeds as follows:

\paragraph{Step 1: Initialization.}
Given the platinum anchor~\texttt{(Pt)}, the algorithm initializes the BM with one column per anchor label present in the training data (here columns for \texttt{(Pt)} and \texttt{(Zr)}) recording how many times each anchor atom appears in each molecule.

\paragraph{Step 2: Path selection and exploration.}
The selector evaluates the BM to identify the most predictive path. As shown in Figure~\ref{fig:path_selection}, at the first iteration the selector identifies \texttt{(Pt,\,C)} as the most informative path. Since this path has not been selected before, the algorithm adds all one-step extensions found in the training graphs as new BM columns. Note that Figure~\ref{tab:bm_ebm} depicts the state at the \emph{second} iteration, after \texttt{(Pt,\,C)} was already selected in iteration~1: the shaded columns \texttt{(Pt,\,C,\,N)} and \texttt{(Pt,\,C,\,C)} are the extensions added upon that selection.

\paragraph{Step 3: EBM expansion.}
Once a path is selected, the EBM is expanded with features for that path and all its prefixes. As shown in the bottom panel of Figure~\ref{tab:bm_ebm}, the EBM for \texttt{(Pt,\,C)} includes the occurrence count and averaged atomic mass of the terminal C atom, together with the averaged bond order of the Pt--C edge; the prefix \texttt{(Pt)} columns were already present from the initialization step.

\paragraph{Step 4: Base learner training.}
The base learner is trained on the EBM columns corresponding to the selected path and its prefixes, fitting to the negative gradient of the current predictions.

\paragraph{Step 5: Repeat.}
Steps 2--4 are repeated for the next boosting iteration, with the updated BM now including the extensions discovered in the previous step. This continues until the specified number of iterations $T$ is reached.

\subsection{Loss functions and boosting details}\label{subsec:boosting_details}

Section~\ref{subsec:boosting} described the generic boosting procedure. Here we
specify how it is instantiated for the two supported tasks, how the package
handles single versus multiple anchor labels, and how the learning rate is
applied externally.

\subsubsection{Regression}\label{subsec:regression}

For regression tasks, the package minimizes the squared error loss:
\begin{equation}
  \ell = \frac{1}{N}\sum_{i=1}^{N}\bigl(y_i - \hat{y}_i\bigr)^2.
\end{equation}
The model is initialized with the mean of the training targets, and pseudo-residuals are the ordinary residuals $r_{t,i} = y_i - \hat{y}_{t-1,i}$. After $T$ boosting iterations, the final prediction is the sum of all base learner outputs, scaled by the learning rate $\eta$.

\subsubsection{Classification}\label{subsec:classification}

For binary classification, the package minimizes the binary cross-entropy loss:
\begin{equation}
  \ell = -\frac{1}{N}\sum_{i=1}^{N}\bigl[y_i \log \hat{p}_i + (1 - y_i)\log(1 - \hat{p}_i)\bigr],
\end{equation}
where $\hat{p}_i$ is the predicted probability of $Y_i=1$ for graph $G_i$. Boosting updates accumulate in log-odds space: the model maintains a score $F(G_i) = \sum_t \eta \cdot h^{(t)}(G_i)$, which is converted to a probability via the sigmoid function $\hat{p}_i = \sigma(F(G_i)) = (1 + e^{-F(G_i)})^{-1}$. The model is initialized with the log-odds of the positive class proportion in the training set. The negative gradient (pseudo-residuals) at each iteration is $r_{t,i} = y_i - \hat{p}_{t-1,i}$, i.e., the difference between the true label and the current predicted probability.

\paragraph{Standard mode.}
By default, the tree is fitted to the pseudo-residuals, and its leaf values are used directly as the boosting update. This is the standard gradient boosting approach and works well in practice for most tasks.

\paragraph{Leaf-optimized mode (\texttt{use\_tree\_boost=True}, tree base learners only).}
When \texttt{use\_tree\_boost=True} is set on \texttt{SequentialPathBoostClassifier}, the package activates the \texttt{BoostedTreeBaselearner}, which follows the TreeBoost approach of \citet{friedman2001greedy}. The tree is still fitted to the pseudo-residuals to determine the partition of graphs into leaves, but the leaf output values are then re-optimized via a line search that directly minimizes the logistic loss in the log-odds space. For each leaf $z$, let $R_z$ denote the set of training graphs assigned to that leaf. The optimal scalar $\gamma_z$ is found by minimizing
\begin{equation}
  \gamma_z = \arg\min_{\gamma} \sum_{G_i \in R_z} \ell\bigl(y_i,\, \sigma(F_{t-1}(G_i) + \gamma)\bigr)
\end{equation}
over $\gamma \in [-10, 10]$. At each boosting iteration, the update for each graph uses the optimized $\gamma_z$ of its leaf in the current base learner, rather than the tree's original leaf value. This approach produces better-calibrated probabilities than the standard mode, since the leaf values are optimized directly for logistic loss rather than squared error with respect to the residuals.

After $T$ boosting iterations, the final class prediction is obtained by thresholding the predicted probability: $\hat{y}_i = \mathbf{1}[\hat{p}_i \geq 0.5]$.

\subsubsection{\texttt{PathBoost} and \texttt{SequentialPathBoost}}\label{subsec:two_algorithms}

The package provides two main classes. \texttt{SequentialPathBoost} implements the algorithm described above for a single anchor label. \texttt{PathBoost} extends this to multiple anchor labels by training independent \texttt{SequentialPathBoost} instances (one per anchor label) and aggregating their predictions. Since paths starting from different anchor labels cannot discover common extensions, these sub-models are independent and can be trained in parallel.
 The choice between the two depends on how many anchor nodes of each label a graph contains. When each graph has exactly one anchor node per label (as in molecular graphs with a unique metal center), the per-label sub-models are fully independent, and \texttt{PathBoost} is recommended. When a graph contains several anchor nodes of the same label, their contributions must be averaged within the sub-model, which can make its output less stable; in this case \texttt{SequentialPathBoost} applied to a single anchor set combining all labels is recommended.

\subsubsection{Learning rate handling}\label{subsec:learning_rate}

The package does not assume that the base learner has an internal learning rate. Instead, the learning rate $\eta$ is applied externally (the default value is 0.1). To ensure that this external scaling behaves correctly, the negative gradient is centered (its mean is subtracted) before fitting the base learner. The mean is then added back after scaling the predictions by $\eta$. This guarantees that the base learner's predictions are centered around zero, so that multiplication by $\eta$ produces a proper shrinkage toward zero.

\subsection{Variable importance}\label{subsec:variable_importance}

The package provides two criteria for computing path importance, both implemented in \texttt{VariableImportance\_ForSequentialPathBoost} and controlled via the \texttt{parameters\_variable\_importance} dictionary. Variable importance is also available for \texttt{PathBoost}, where per-anchor importances are combined into a single global score as described in the cross-anchor combination paragraph below. Both criteria can optionally apply a correlation adjustment and normalization. The error metric used (\texttt{'mse'} or \texttt{'mae'}) is set via the \texttt{'error\_used'} key (this controls importance scoring only, it does not affect how base learners are fitted).

\paragraph{Absolute importance (default).}
A path's importance is the sum of all the improvements in training error over the iterations in which the path was selected. At each boosting iteration $t \geq 1$, the error improvement is
\begin{equation}
  \Delta_t = \ell_{t-1} - \ell_t,
\end{equation}
where $\ell_t$ is the training loss after iteration $t$. This improvement is accumulated for each path $p$ selected at iteration $t$:
\begin{equation}
  \mathrm{importance}(p) = \sum_{t\,:\,\mathrm{selected}(t)=p} \Delta_t.
\end{equation}
As a special case, the path selected at iteration $0$ is assigned the improvement with respect to the mean of the target variables $y$.

\paragraph{Relative importance.}
Relative importance measures, at each iteration, how much better the selected
path is compared to the best available alternative. This complements absolute
importance: a path can accumulate high absolute importance simply because it is
selected early and often, even when other paths would have reduced the error
almost as much. Relative importance instead rewards paths that are
\emph{irreplaceable},those for which no comparably predictive alternative was
available at the time of selection, which is especially informative when many
candidate paths are correlated (for example nested paths sharing a common
prefix) and one wants to identify the substructures that uniquely drive
predictions. At iteration $t$, the frequency column of the selected path $p^*_t$ is temporarily removed from the BM, the selector is re-run on the remaining columns to identify the second-best path $p^{(2)}_t$, and an additional base learner is fitted on $p^{(2)}_t$. The importance contribution is the error gap
\begin{equation}
  \mathrm{contribution}_t = \ell^{(2)}_t - \ell_t,
\end{equation}
where $\ell^{(2)}_t$ is the training error that would have resulted from selecting $p^{(2)}_t$ instead. If a path has high relative importance it means it is substantially better than any available alternative at the time of selection. Because the selector operates on the BM while the base learner operates on the full EBM, it is possible (though rare) for the second-best path to lead to a lower error, producing a negative contribution. Relative importance requires re-running the selector and fitting an additional base learner at every iteration, making it significantly more expensive than absolute importance. In the Python function, it must be enabled explicitly by setting \texttt{'criterion': 'relative'}.

\paragraph{Correlation adjustment.}
Because longer paths extend shorter ones as a prefix, their occurrence in a graph is correlated by construction. This means that if the shorter path drove most of the error reduction early in training, the longer extension may appear unimportant even though its presence is linked to the shorter path. When \texttt{'use\_correlation': True} is set, the importance of each path $p$ is given by the sum of correlation-weighted importances of all its prefixes (see Section~\ref{sec:terminology}):
\begin{equation}
  \mathrm{importance}_{\mathrm{corr}}(p)
    = \mathrm{importance}(p) + \sum_{\substack{q:\, |q| \leq |p| \\ q_i = p_i,\, \forall i \leq |q|}} \rho(p,\,q)\cdot\mathrm{importance}(q),
\end{equation}
where $\rho(p, q)$ is the Pearson correlation between the occurrence counts of $p$ and $q$ across all training graphs.

\paragraph{Normalization.}
When \texttt{'normalize': True} (the default), importances are rescaled to sum to \texttt{normalization\_value} (default: 100):
\begin{equation}
  \mathrm{importance}_{\mathrm{norm}}(p)
    = \frac{\mathrm{importance}(p)}{\sum_{p'} \mathrm{importance}(p')} \times 100.
\end{equation}
For the absolute criterion without correlation adjustment, this makes importances directly interpretable as percentage contributions to total explained error. For other criteria, normalization provides a common scale for comparing path importances across runs, but the percentage-of-error interpretation is not as straightforward.

\paragraph{Cross-anchor combination.}
When \texttt{PathBoost} trains one \texttt{SequentialPathBoost} model per anchor label, the per-anchor importances are combined via a weighted average, where the weight for anchor $a$ is proportional to the number of training graphs $n_a$ in which that anchor appears. The combined importances are then re-normalized to \texttt{normalization\_value}. The final importance scores are stored in \texttt{model.variable\_importance\_}, mapping each selected path (e.g., \texttt{(Mn, N, O)}) to its scalar importance.

\section{Software design and implementation} \label{sec:implementation}

This section describes the practical aspects of using the \texttt{path\_boost} package.

\subsection{Data requirements}\label{subsec:data_requirements}

\paragraph{Attribute types.}
All attributes are treated uniformly: any attribute whose values are numbers (instances of \texttt{numbers.Number}) is added to the EBM, while attributes with non-numeric values are ignored. This means categorical attributes that the user wants to include must be encoded numerically beforehand, for example via one-hot encoding or an integer representation. This applies to the anchor attribute as well, which both defines the paths and contributes its own averaged value as a feature. Finally if a node and an edge share an attribute name, the edge attribute takes precedence.

\paragraph{Graph types.}
The package supports directed and undirected graphs from \texttt{networkx}. In
directed graphs, edge orientation matters both when a path is extended and when paths are compared: two paths are considered identical only if they visit the same node labels and agree on the orientation of every edge. For example, $a \rightarrow b \rightarrow c$ and $a \rightarrow b \leftarrow c$ are distinct paths.
Multigraphs are also accepted,
but in this case each edge is treated as bidirectional.

\subsection{Multiple paths handling}\label{subsec:multiple_paths}

When multiple instances of the same labeled path exist in a graph, their attribute values are aggregated by averaging over all occurrences. To use a different aggregation function (e.g., sum or max), the the user can create a subclass of EBM and override the static method \texttt{combine\_attribute\_value\_of\_multiple\_paths\_in\_the\_same\_graph}.

\subsection{Extended Boosting Matrix structure}\label{subsec:ebm_structure}

The EBM is a matrix where rows correspond to individual graphs and columns correspond to path-based features, following the naming convention \texttt{(path)\_attribute}. For example, a column named \texttt{(6, 7, 8)\_atomic\_mass} contains the averaged atomic mass of oxygen atoms (atomic number 8) reached via the path carbon$\rightarrow$nitrogen$\rightarrow$oxygen. After fitting, the full EBM can be inspected via the \texttt{train\_ebm\_dataframe\_} attribute:

\begin{lstlisting}
>>> # Example column names from a trained model
>>> print(model.train_ebm_dataframe_.columns[:5])
Index(['(25,)_n_times_present', '(25,)_atomic_number',
   '(25,)_n_hydrogens', '(25, 7)_n_times_present',
   '(25, 7)_bond_strength'], dtype='object')
\end{lstlisting}

\subsection{Path length and feature-space growth}

The parameter with the largest impact on the cost of training is \texttt{max\_path\_length}. The number of distinct labelled paths that can be discovered grows combinatorially with both the path length and the diversity of node labels: at each additional step a path may branch into as many extensions as there are distinct neighbouring labels, so the space of candidate paths grows roughly exponentially in \texttt{max\_path\_length} and in the branching factor of the label set. Because every discovered path contributes several columns to the Extended Boosting Matrix (EBM, one per node and edge attribute of the path and its prefixes, the width of the EBM, and hence the memory and time needed to build and fit it, scales with the number of paths actually realized during training.

Two mechanisms keep this growth manageable. First, paths are expanded \emph{lazily}: the boosting matrix is extended only when a path is selected, so the realized feature space is far smaller than the full combinatorial enumeration and is concentrated on structurally informative paths (Section~\ref{subsec:boosting}). Second, \texttt{max\_path\_length} caps the depth of exploration directly. The realized EBM can be inspected after fitting through the \texttt{train\_ebm\_dataframe\_} attribute (Section~\ref{subsec:ebm_structure}), which is useful for diagnosing how large the feature space has actually become on a given dataset.

In practice we recommend treating \texttt{max\_path\_length} as a primary
hyperparameter and tuning it by cross-validation, starting from small values
(e.g.\ 3--4) and increasing until the validation error stops improving. Short
paths may miss long-range structural signal, whereas overly long paths inflate
the EBM, raise training time and memory, and add many weakly informative
features that increase the risk of overfitting. The appropriate value is
dataset-dependent: graphs with diverse node labels (for example molecular
datasets spanning many chemical elements) branch more quickly and reach a large
EBM at smaller path lengths than graphs with few distinct labels.

\subsection{Selector}\label{subsec:custom_selector}

The default selector for both regression and classification is
\texttt{sklearn.tree.DecisionTreeRegressor} with \texttt{max\_depth=1} (a
decision stump), passed via the \texttt{SelectorClass} constructor parameter.

The selector's sole purpose is path selection: at each boosting iteration it
identifies which candidate path in the BM is most predictive of the current
negative gradient. Concretely, it is fitted on the frequency-only columns of
the BM (one column per known path, recording how many times that path appears
in each graph) and the path corresponding to the highest entry in
\texttt{feature\_importances\_} is selected. The selector is never called
during prediction. This design is intentional: keeping the selector focused on
structural frequency counts rather than attribute values makes path selection
fast and robust to overfitting, as discussed in
Section~\ref{subsec:boosting}.

\paragraph{Custom selectors.}
The selector interface is duck-typed via \texttt{SelectorClassInterface}:
explicit inheritance is not required, and any class implementing the expected
methods is automatically recognized as compatible. A custom selector must
implement \texttt{fit(X, y)}, \texttt{predict(X)} (required by the interface
but never called during training), and expose a \texttt{feature\_importances\_}
attribute after fitting, namely an array of non-negative scores, one per column
of \texttt{X}. Only the ordering of scores matters, not their scale. In principle, any
\texttt{scikit-learn} estimator that exposes \texttt{feature\_importances\_}
can be used directly as a selector. 

\subsection{Available base learners}\label{subsec:base_learners}

The default base learner for both \texttt{SequentialPathBoost} and
\texttt{PathBoost} (see Section~\ref{subsec:two_algorithms}) is
\texttt{sklearn.tree.DecisionTreeRegressor} with \texttt{max\_depth=3}, passed
via the \texttt{BaseLearnerClass} constructor parameter. Shallow decision trees
handle NaN values natively, which arise in the EBM when a path is absent from a
graph, and are well suited to the small number of columns (typically 3--10)
that the base learner receives at each iteration, corresponding to the EBM
features of the selected path and its prefixes.

\paragraph{Custom base learners.}
As with the selector, base learners use a duck-typed interface, defined in
\texttt{BaseLearnerClassInterface}: explicit inheritance is not required, and
any class implementing \texttt{fit(X, y)} and \texttt{predict(X)} is
automatically recognized as compatible. The \texttt{X} argument is always a
\texttt{pd.DataFrame} containing only the EBM columns for the currently
selected path; it is not the full feature matrix. Any \texttt{scikit-learn}
estimator with \texttt{fit} and \texttt{predict} (e.g., \texttt{LinearRegression},
\texttt{XGBRegressor}) works without modification. Table~\ref{tab:base_learners}
summarizes the base learners that have been tested for compatibility with the package.

\begin{table}[htbp]
\caption{Base learner compatibility. Tested learners have been verified in the package test suite.}
\label{tab:base_learners}
\begin{tabular}{lll}
\toprule
Library & Class & Notes \\
\midrule
scikit-learn & \texttt{DecisionTreeRegressor} & Default (\texttt{max\_depth=3}) \\
scikit-learn & \texttt{LinearRegression} & Requires \texttt{replace\_nan\_with} \\
XGBoost & \texttt{XGBRegressor} & Also usable as selector \\
scikit-learn & \texttt{Ridge}, \texttt{Lasso}, \texttt{ElasticNet} & Requires \texttt{replace\_nan\_with} \\
scikit-learn & \texttt{Pipeline(SplineTransformer, Ridge)} & Requires \texttt{replace\_nan\_with}; smooth targets \\
scikit-learn & \texttt{SVR} & Cannot be used as selector \\
\bottomrule
\end{tabular}
\end{table}

\paragraph{NaN handling.}
By default, EBM entries are \texttt{np.nan} when a path is absent from a graph.
Tree-based models handle this natively. Linear models do not: when using
\texttt{LinearRegression}, \texttt{Ridge}, or similar, the
\texttt{replace\_nan\_with} parameter must be set to a numeric value:
\begin{lstlisting}
path_boost = PathBoost(
    BaseLearnerClass=LinearRegression,
    replace_nan_with=0,
    ...
)
\end{lstlisting}
The choice of replacement value matters: \texttt{0} is a neutral default, while
a large negative value such as \texttt{-100} signals path absence explicitly to
the model. For most use cases \texttt{0} is the safer choice, unless missing
paths need to be distinguished from paths with a true zero attribute value.

\paragraph{Classification variant.}
The \texttt{SequentialPathBoostClassifier} class provides an additional option
via the \texttt{use\_tree\_boost} flag, which activates an internal base learner
that optimizes per-leaf values by minimizing the logistic loss directly,
following the approach of \citet{friedman2001greedy}:
\begin{lstlisting}
from path_boost import SequentialPathBoostClassifier
model = SequentialPathBoostClassifier(
    use_tree_boost=True,
    ...
)
\end{lstlisting}
When \texttt{use\_tree\_boost=True}, the \texttt{BaseLearnerClass} parameter is
ignored.

\subsection{Parallelization}\label{subsec:parallelization}

The \texttt{PathBoost} class supports parallel training across anchor labels by setting \texttt{n\_of\_cores > 1}. Thread limits are configured at module load time to prevent over-subscription. For cross-platform compatibility, the main script should be wrapped with:
\begin{lstlisting}
if __name__ == "__main__":
    path_boost = PathBoost(n_of_cores=4)
    path_boost.fit(X, y, ...)
\end{lstlisting}

\subsection{Model persistence}\label{subsec:persistence}

Trained models can be saved and loaded for reproducibility:
\begin{lstlisting}
>>> path_boost.save('trained_model.joblib')
>>> from path_boost import PathBoost
>>> loaded_model = PathBoost.load('trained_model.joblib')
>>> predictions = loaded_model.predict(X_test)
\end{lstlisting}


\subsection{Tuning the number of iterations}\label{subsec:cross_validation}

As anticipated in Section~\ref{subsec:boosting}, the number of boosting iterations $T$ is one of the most influential hyperparameters of the model. The package offers two complementary ways of choosing it: \emph{early stopping} on a held-out evaluation set, and \emph{cross-validation} over a grid of candidate values. This section describes each mechanism in turn and explains how they should (and should not) be combined.

\paragraph{Early stopping and the \texttt{patience} parameter.} Early stopping requires an evaluation set, passed to \texttt{fit()} through the \texttt{eval\_set} argument. After every boosting iteration, the model predicts on this set, records the evaluation MSE, and checks whether the active stopping criterion has been triggered. The \texttt{patience} parameter is an integer that controls how tolerant the procedure is to short-term fluctuations in the evaluation curve: training is halted once the evaluation MSE has failed to improve on its best value for \texttt{patience} consecutive iterations. This tolerance is needed because the evaluation MSE does not always decrease monotonically. A small \texttt{patience} (e.g.\ 5) reacts quickly to real overfitting, but it can also stop on one of these temporary upticks and treat it as overfitting when it is not, halting in a local minimum of the evaluation curve before training reaches a deeper minimum a few iterations later. A larger value (e.g.\ 20--50) ignores these small fluctuations and is more likely to find the best model, at the cost of more iterations. Stopping too early is partly offset by \texttt{restore\_best\_model} (discussed below), which rolls the model back to the lowest-MSE checkpoint; \texttt{patience} still controls how far past the current best the search keeps going. The alternative criterion, \texttt{target\_error}, halts training as soon as the evaluation MSE drops to or below a user-specified threshold; it is useful when a problem-specific error target is known a priori. Without an \texttt{eval\_set} both criteria are silently ignored and training always runs for the full \texttt{n\_iter} iterations. When \texttt{restore\_best\_model=True} (the default), after stopping the model is rolled back to the checkpoint with the lowest evaluation MSE, so the returned estimator is not the (possibly overfitted) one from the final iteration. To be able to use \texttt{restore\_best\_model} an evaluation dataset (typically a holdout of the training data) must be passed. Choosing the best checkpoint on it is therefore a model-selection step on validation data, just as in ordinary early stopping for gradient boosting or neural networks. Because this set decides when to stop, its MSE is an optimistic measure of accuracy and should not be reported as the final result: that must come from a separate test set that took no part in training or early stopping, as done in the tutorial (Section~\ref{sec:application}) and benchmark (Appendix~\ref{app:benchmark}).

\paragraph{Cross-validation.} Cross-validation is supported through the standard \texttt{sklearn} interface: \texttt{PathBoost} can be passed directly to utilities such as \texttt{GridSearchCV} together with a grid over the hyperparameters of interest (typically \texttt{learning\_rate}, \texttt{max\_path\_length}, and, when desired, \texttt{n\_iter}). For each candidate configuration the model is refit on every fold and the validation score is averaged across folds, yielding a fair estimate of generalisation error.

\paragraph{Combining the two strategies.} Early stopping and cross-validation should not be combined in a single \texttt{GridSearchCV} call: forwarding an \texttt{eval\_set} through \texttt{fit\_params} reuses the same validation set across every fold, and different folds stop at different iterations, so the cross-validated score no longer corresponds to a single, well-defined model. We recommend keeping \texttt{n\_iter} fixed during cross-validation and applying early stopping only in the final refit.

\section{Tutorial} \label{sec:application}

\subsection{Installation}

The package can be installed from the Python Package Index via:
\begin{lstlisting}
pip install path-boost
\end{lstlisting}
Dependencies can be installed with:
\begin{lstlisting}
pip install -r requirements.txt
\end{lstlisting}

\subsection{Molecular property prediction on the tmQMg dataset}

We demonstrate the package on the tmQMg dataset \citep{D2DD00129B}, a collection of transition metal compounds represented as molecular graphs. In this dataset, atoms are nodes and bonds are edges. Each node carries numerical attributes derived from the periodic table (e.g., atomic number, atomic mass), and edges carry bond-related attributes (e.g., bond strength). The target variable is the HOMO--LUMO gap, a continuous molecular property. The atomic number serves as the anchor attribute.

We initialize \texttt{PathBoost} with the parameters that will remain constant during cross-validation, for example the fixed number of iterations \texttt{n\_iter}, together with the variable-importance settings. The grid search will then vary only \texttt{learning\_rate}, \texttt{max\_path\_length}, and \texttt{kwargs\_for\_base\_learner}. A full description of each parameter is available in Appendix~\ref{app:technical}.

\begin{lstlisting}
vi_params: dict = {
    'criterion': 'absolute',   # 'absolute' or 'relative'
    'error_used': 'mse',       # metric for importance scoring only, not for training
    'use_correlation': False,
    'normalize': True,
    }
path_boost = PathBoost(
    n_iter=1000, # max number of iterations
    max_path_length=6,
    learning_rate=0.02,
    n_of_cores=1,
    verbose=True,
    parameters_variable_importance=vi_params
)
\end{lstlisting}
Note that \texttt{error\_used} controls only how variable importance scores are computed after fitting (MSE vs.\ MAE when measuring a path's contribution to error reduction); it has no effect on how the base learners are trained.

To perform cross validation we use the standard sklearn procedure.
\begin{lstlisting}
# Define the parameter grid
param_grid = {
    'learning_rate': [0.01, 0.8],
    'max_path_length': [3 ,4],
    'kwargs_for_base_learner': [{'max_depth': 3}, {'max_depth': 5}]
    }

# Initialize GridSearchCV
grid_search = GridSearchCV(
    estimator=path_boost,
    param_grid=param_grid,
    cv=2,
    scoring='neg_mean_squared_error',
    verbose=3
)

\end{lstlisting}

A compact, self-contained version of the same workflow, without the surrounding commentary, is provided in Appendix~\ref{app:workflow} for quick reference.

As discussed in Section~\ref{subsec:cross_validation}, we recommend leaving \texttt{patience} and \texttt{target\_error} unset during \texttt{GridSearchCV}, so that each fold is trained for the same fixed number of iterations and the cross-validated scores remain directly comparable. Early stopping is reintroduced in the final training run below.

\subsection{Training}
\begin{figure}[t!]
    \centering
    \begin{subfigure}{0.48\textwidth}
        \centering
        \includegraphics[width=\linewidth]{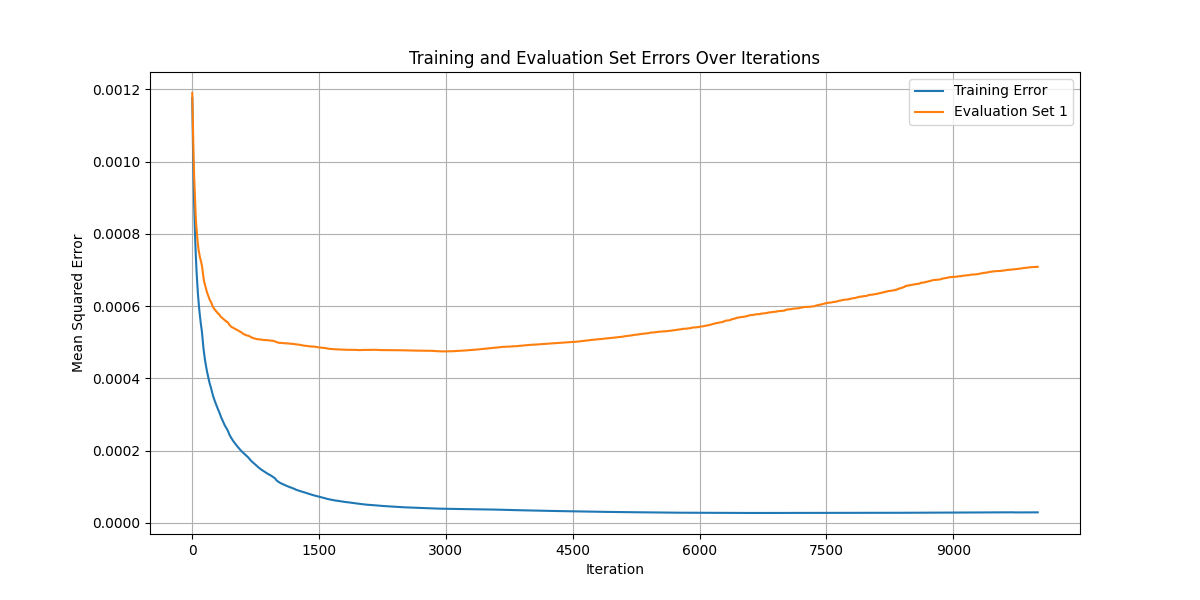}
        \caption{Training and validation MSE across boosting iterations. The model converges within 100 iterations.}
        \label{fig:train_and_eval_error}
    \end{subfigure}
    \hfill
    \begin{subfigure}{0.48\textwidth}
        \centering
        \includegraphics[width=\linewidth]{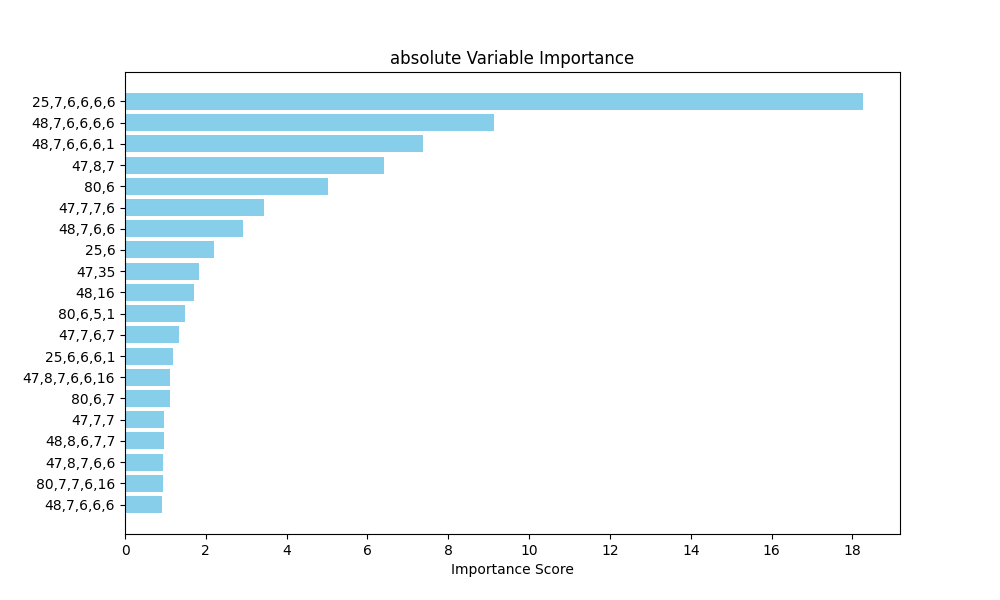}
        \caption{Top 15 paths by importance. Path labels correspond to atomic numbers (e.g., 25=Mn, 7=N, 8=O).}
        \label{fig:absolute_variable_importance}
    \end{subfigure}
    \caption{PathBoost performance on the tmQMg dataset: (a) MSE across iterations shows convergence and overfitting, (b) Variable importance reveals which labeled paths contribute most to HOMO--LUMO gap predictions.}
    \label{fig:path_boost_performances}
\end{figure}

For the final training run, the model should be re-initialized with the best hyperparameters and a \texttt{patience} value for early stopping. Here \texttt{n\_iter} serves as the maximum number of iterations for each sub-model. This is the strategy we adopt here, but it is not the only one: a common alternative is to fix the other hyperparameters in advance and instead cross-validate the number of iterations directly (Section~\ref{subsec:cross_validation}).
\begin{lstlisting}
path_boost = PathBoost(
    n_iter=1000,
    patience=5,
    max_path_length=6,
    learning_rate=0.02,
    n_of_cores=1,
    verbose=True,
    parameters_variable_importance=vi_params
)

path_boost.fit(
    X=X_train,
    y=y_train,
    eval_set=eval_set,
    list_anchor_nodes_labels=anchor_labels,
    anchor_nodes_label_name="feature_atomic_number",
    ...
)
\end{lstlisting}
The best hyperparameters from cross-validation should be included in the initialization. After training, the model's performance and feature importance can be inspected:
\begin{lstlisting}
path_boost.plot_training_and_eval_errors(skip_first_n_iterations=False)

path_boost.plot_variable_importance(top_n_features=15)
\end{lstlisting}
The result is shown in Figure \ref{fig:path_boost_performances}

\subsection{Prediction}
To perform a prediction, input a list of \texttt{nx.Graph} objects:

\begin{lstlisting}
predictions = path_boost.predict(X_test)
\end{lstlisting}
When working directly with \texttt{SequentialPathBoost}, the EBM can be pre-computed once and reused across multiple calls to avoid recomputation:
\begin{lstlisting}
ebm = seq_path_boost.generate_ebm_for_dataset(X_test)
predictions = seq_path_boost.predict(ebm_dataframe=ebm)
\end{lstlisting}
Note that \texttt{PathBoost.predict()} does not expose this shortcut; it always recomputes the EBM internally.

\section{Benchmark comparison}\label{sec:graphbench}

We compare PathBoost against two established baselines across six molecular regression tasks. The baselines are the Graph Isomorphism Network with Edge features (GINE) \citep{hu2020strategies} and the Weisfeiler--Leman graph kernel paired with support vector regression (WL~+~SVR) \citep{Kriege2020}. All methods are evaluated using 5-fold cross-validation repeated three times, yielding 15 train/test evaluations per method and dataset. Within each fold the training set is further split 90/10 into a training and a validation subset used for early stopping and, where applicable, hyperparameter selection; the validation subset is never used for final evaluation. Reported metrics (MAE, RMSE, R$^2$) are the mean and standard deviation across the 15 folds, together with the average training time per fold. All experiments were run on CPU.

PathBoost was applied with fixed hyperparameters, with anchor labels automatically extracted from each training fold as the set of unique atomic numbers present in that fold. The two baselines were instead tuned by grid search on the validation subset of each fold (minimising validation MAE) and then refit on the combined train and validation set using the best configuration. The full hyperparameter grids and fixed settings for all three methods are reported in Appendix~\ref{app:benchmark}. This benchmark focuses on regression tasks; a comparative study that also covers classification on a broader range of graph datasets is presented in \citet{meggio_2nd}.

\subsection{Datasets}

\paragraph{ESOL.}
ESOL \citep{wu2018moleculenet} is a small dataset of 1{,}128 organic compounds with experimentally measured water solubility expressed as log(mol/L). It is part of the MoleculeNet benchmark collection \citep{wu2018moleculenet}.

\paragraph{FreeSolv.}
FreeSolv \citep{wu2018moleculenet} provides experimental hydration free energies (kcal/mol) for 643 small molecules in water, derived from alchemical free energy calculations.

\paragraph{QM9.}
QM9 \citep{ramakrishnan2014quantum} contains quantum-mechanical properties of approximately 134k stable small organic molecules composed of C, H, O, N, and F atoms with up to 9 heavy atoms, computed at the B3LYP/6-31G(2df,p) level of theory. The target property is the HOMO--LUMO gap. Due to its size, we use a random subsample of 10{,}000 graphs.

\paragraph{tmQMg.}
The tmQMg dataset \citep{D2DD00129B} is a collection of mononuclear transition metal complexes extracted from the Cambridge Structural Database, represented as molecular graphs where atoms are nodes and bonds are edges.From the u-NatQG variant we consider three regression targets, each treated as a separate task:
\begin{itemize}
  \item dipole moment;
  \item HOMO energy;
  \item polarizability.
\end{itemize}
Due to the size of the dataset, each target uses a random subsample of 5{,}000 graphs.

\subsection{Results}

Table~\ref{tab:comparison} reports the benchmark results. PathBoost achieves the best performance on five of the six tasks. On the two small organic datasets, ESOL and FreeSolv, PathBoost outperforms both baselines by a clear margin across all metrics, suggesting that path-based features are particularly effective when training data is limited. On QM9, GINE performs best, which is consistent with the known strength of graph neural networks on large, homogeneous organic datasets where message-passing can exploit rich local chemical environments. On the three tmQMg targets, PathBoost performs best on all the considered tasks, with especially strong results on polarizability ($R^2 = 0.928$) and HOMO energy ($R^2 = 0.584$), while all methods struggle on the dipole moment target, reflecting the intrinsic difficulty of predicting this property from graph topology alone.

Regarding computational cost, GINE is substantially slower than both PathBoost and WL~+~SVR across all datasets, with training times up to an order of magnitude larger on the tmQMg tasks. Note that all methods were run on CPU; GPU acceleration would be expected to reduce GINE's training time considerably. PathBoost occupies a middle ground, being faster than GINE while achieving better predictive performance on most tasks.

\begin{table}[htbp]
\caption{Regression performance comparison. Mean $\pm$ standard deviation
over 5-fold cross-validation repeated 3 times. Best results per dataset
are shown in bold.}
\label{tab:comparison}
\centering
{\small\setlength{\tabcolsep}{4pt}%
\begin{tabular}{llcccc}
\toprule
  \textbf{Dataset} & \textbf{Method} & \textbf{MAE} & \textbf{RMSE} & \textbf{R$^2$} & \textbf{Time (s)} \\
\midrule
  ESOL & PathBoost & \textbf{0.5536 $\pm$ 0.0257} & \textbf{0.7345 $\pm$ 0.0344} & \textbf{0.8759 $\pm$ 0.0121} & 11.0 \\
   & GINE & 0.6922 $\pm$ 0.0494 & 0.9446 $\pm$ 0.0750 & 0.7941 $\pm$ 0.0328 & 64.4 \\
   & WL~+~SVR & 0.7096 $\pm$ 0.0451 & 0.9576 $\pm$ 0.0768 & 0.7880 $\pm$ 0.0354 & 1.0 \\
  \midrule
  FreeSolv & PathBoost & \textbf{0.7857 $\pm$ 0.0954} & \textbf{1.1809 $\pm$ 0.1832} & \textbf{0.9027 $\pm$ 0.0267} & 5.2 \\
   & GINE & 1.0027 $\pm$ 0.1024 & 1.4333 $\pm$ 0.1858 & 0.8573 $\pm$ 0.0331 & 27.1 \\
   & WL~+~SVR & 1.0350 $\pm$ 0.0836 & 1.4885 $\pm$ 0.1687 & 0.8476 $\pm$ 0.0243 & 0.4 \\
  \midrule
  QM9 & PathBoost & 0.0227 $\pm$ 0.0019 & 0.0284 $\pm$ 0.0019 & 0.6429 $\pm$ 0.0480 & 13.2 \\
   & GINE & \textbf{0.0122 $\pm$ 0.0007} & \textbf{0.0185 $\pm$ 0.0013} & \textbf{0.8494 $\pm$ 0.0208} & 425.8 \\
   & WL~+~SVR & 0.0285 $\pm$ 0.0005 & 0.0366 $\pm$ 0.0006 & 0.4095 $\pm$ 0.0193 & 57.8 \\
  \midrule
  tmQMg (dipole) & PathBoost & \textbf{2.3558 $\pm$ 0.1277} & \textbf{3.1206 $\pm$ 0.1752} & \textbf{0.3819 $\pm$ 0.0558} & 530.3 \\
   & GINE & 2.6581 $\pm$ 0.0910 & 3.5125 $\pm$ 0.1321 & 0.2169 $\pm$ 0.0508 & 921.4 \\
   & WL~+~SVR & 2.5505 $\pm$ 0.0730 & 3.4264 $\pm$ 0.1524 & 0.2547 $\pm$ 0.0575 & 18.7 \\
  \midrule
  tmQMg (HOMO) & PathBoost & \textbf{0.0235 $\pm$ 0.0017} & \textbf{0.0340 $\pm$ 0.0029} & \textbf{0.5841 $\pm$ 0.0650} & 107.6 \\
   & GINE & 0.0349 $\pm$ 0.0018 & 0.0501 $\pm$ 0.0025 & 0.0984 $\pm$ 0.0778 & 550.6 \\
   & WL~+~SVR & 0.0310 $\pm$ 0.0010 & 0.0487 $\pm$ 0.0019 & 0.1502 $\pm$ 0.0404 & 18.8 \\
  \midrule
  tmQMg (pol.) & PathBoost & \textbf{20.5218 $\pm$ 0.9969} & \textbf{30.9270 $\pm$ 3.8183} & \textbf{0.9284 $\pm$ 0.0153} & 456.7 \\
   & GINE & 39.5176 $\pm$ 1.7912 & 56.2039 $\pm$ 3.1849 & 0.7649 $\pm$ 0.0195 & 1036.3 \\
   & WL~+~SVR & 39.2982 $\pm$ 1.4333 & 56.6162 $\pm$ 3.5413 & 0.7613 $\pm$ 0.0236 & 17.9 \\
\bottomrule
\end{tabular}%
}
\end{table}


\section{Summary and discussion} \label{sec:summary}

PathBoost, introduced in \citet{meggio2024pathboost} and \citet{meggio_2nd}, is a gradient boosting framework for learning from graph-structured data. Its distinguishing feature is interpretability: predictions are decomposed into contributions from labelled paths through the input graphs, and the importance of each path can be quantified after training. While previous works introduced the method and released code for their own experiments, they did not provide a general-purpose easy-to-use implementation.

The \texttt{path\_boost} package presented in this paper fills this gap. It provides a tested, open-source, \texttt{scikit-learn}-compatible reference implementation supporting both regression and binary classification, multiple anchor labels, parallel training across sub-models, and early stopping. Two variable importance criteria, absolute and relative with an optional correlation adjustment, let users identify which labelled paths drive the model's predictions; this matters in scientific applications such as computational chemistry, where understanding \emph{why} a prediction was made is often as important as the prediction itself. By following \texttt{scikit-learn} conventions and using \texttt{networkx} for graph handling, the package also integrates naturally with existing data science workflows, enabling cross-validation via \texttt{GridSearchCV} and custom scores, without requiring any custom plumbing on the user's side.



\section*{Computational details}

The results in this paper were obtained using Python~3.12 with the following key packages: \texttt{path\_boost}~2.1.0, \texttt{scikit-learn}~1.6.1 \citep{pedregosa2011scikit}, \texttt{networkx}~3.4.2 \citep{hagberg2008exploring}, \texttt{pandas}~2.2.3 \citep{mckinney2010data}, \texttt{numpy}~2.1.3 \citep{harris2020array}, and \texttt{matplotlib}~3.9.3 \citep{hunter2007matplotlib}.

\backmatter

\section*{Statements and Declarations}

\paragraph{Funding.}
The authors disclosed receipt of the following financial support for the research, authorship and/or publication of this article: CM: CompSci (EU Horizon 2020 MSCA, n. 945371). JP: Integreat (RCN, n. 332645). RDB: Integreat (RCN, n. 332645), Plumbin’ (RCN, n. 323985).

\paragraph{Competing interests.}
The authors have no competing interests to declare that are relevant to the content of this article.

\paragraph{Ethics approval and consent to participate.}
Not applicable.

\paragraph{Data availability.}
All datasets used in this work are publicly available. ESOL and FreeSolv are distributed as part of the MoleculeNet benchmark collection \citep{wu2018moleculenet}. QM9 is available from \citet{ramakrishnan2014quantum}. The tmQMg dataset is described in \citet{D2DD00129B} and the version used here is available at \url{https://github.com/uiocompcat/tmQMg}.

\paragraph{Code availability.}
The \texttt{path\_boost} package is open-source and freely available. The source code is hosted on GitHub at \url{https://github.com/Claudio-Me/extended_path_boost}, and releases are distributed on the Python Package Index (PyPI) at \url{https://pypi.org/project/path-boost/}. The version used to produce the results in this paper is reported in the Computational details section.

\begin{appendices}






\section{API reference} \label{app:technical}

This appendix provides a summary of the main classes and their parameters.

\subsection{PathBoost Parameters}

\begin{itemize}
  \item \texttt{n\_iter}: Maximum number of boosting iterations (default: 100)
  \item \texttt{max\_path\_length}: Maximum length of discovered paths (default: 5)
  \item \texttt{learning\_rate}: Shrinkage parameter for boosting updates (default: 0.1)
  \item \texttt{n\_of\_cores}: Number of CPU cores for parallel training (default: 1)
  \item \texttt{patience}: Early stopping patience; halts training if validation loss does not improve for this many consecutive iterations (default: None)
  \item \texttt{restore\_best\_model}: If \texttt{True}, the model is restored to the checkpoint with the best validation performance after early stopping, rather than retaining the weights from the final iteration (default: True)
  \item \texttt{verbose}: Whether to print progress information (default: False)
\end{itemize}

\subsection{SequentialPathBoost Parameters}

The \texttt{SequentialPathBoost} class accepts the same parameters as \texttt{PathBoost}, except for \texttt{n\_of\_cores} which is not applicable to single-anchor training.

\section{Example workflow} \label{app:workflow}

A complete workflow for using PathBoost on molecular data:

\begin{lstlisting}
from path_boost import PathBoost
import networkx as nx

# 1. Load your graph data and split into train / validation / test
#    sets (each a list of nx.Graph objects)
# 2. Define anchor nodes (e.g., metal centers)
anchor_labels = [25, 47, 48, 80]  # Mn, Ag, Cd, Hg

# 3. Initialize and train the model
model = PathBoost(
    n_iter=100,
    max_path_length=6,
    learning_rate=0.05,
    patience=20,
    verbose=True
    )

model.fit(
    X=train_graphs,
    y=train_targets,
    eval_set=[(val_graphs, val_targets)], # validation holdout, not the test set
    list_anchor_nodes_labels=anchor_labels,
    anchor_nodes_label_name="feature_atomic_number"
    )

# 4. Make predictions on the held-out test set
predictions = model.predict(test_graphs)

# 5. Analyze feature importance
print(model.variable_importance_)

# 6. Save the trained model
model.save('my_model.joblib')
\end{lstlisting}

\section{Benchmark experimental protocol} \label{app:benchmark}

\subsection{PathBoost hyperparameters}

PathBoost was run with fixed hyperparameters without grid search. Anchor labels are automatically extracted from each training fold as the set of unique atomic numbers present. The parameters used are listed in Table~\ref{tab:epb_params}.

\begin{table}[htbp]
\caption{PathBoost hyperparameters used in the benchmark.}
\label{tab:epb_params}
\centering
\begin{tabular}{ll}
\toprule
\textbf{Parameter} & \textbf{Value} \\
\midrule
\texttt{n\_iter} & 500 \\
\texttt{max\_path\_length} & 5 \\
\texttt{learning\_rate} & 0.05 \\
\texttt{patience} & 20 \\
\texttt{restore\_best\_model} & True \\
\bottomrule
\end{tabular}
\end{table}

\subsection{WL Kernel + SVR hyperparameters}

The number of WL iterations and the SVR regularisation parameter $C$ were selected by grid search minimising validation MAE, and then the model was refit on the combined train and validation set using the best configuration. The target $y$ was standardised with \texttt{StandardScaler}. The search grid and fixed settings are listed in Table~\ref{tab:wl_params}.

\begin{table}[htbp]
\caption{WL~+~SVR hyperparameter grid. SVR $\varepsilon = 0.1$ and kernel normalisation were fixed throughout.}
\label{tab:wl_params}
\centering
\begin{tabular}{ll}
\toprule
\textbf{Parameter} & \textbf{Values} \\
\midrule
WL iterations & \{1, 2, 3, 4, 5\} \\
SVR $C$ & \{0.01, 0.1, 1, 10, 100\} \\
\bottomrule
\end{tabular}
\end{table}

\subsection{GINE hyperparameters}

Six GINE configurations (combinations of \texttt{hidden\_dim} and \texttt{num\_layers}) were evaluated by grid search minimising validation MAE; the best configuration was then refit on the combined train and validation set with early stopping. Node features are one-hot encoded atomic numbers (11 classes: H, C, N, O, F, P, S, Cl, Br, I, and other); edge features are bond type encoded as a scalar. The architecture consists of a linear embedding layer followed by \texttt{num\_layers} GINEConv blocks (with batch normalisation and ReLU activations) and a global mean pooling layer feeding an MLP prediction head. The hyperparameter grid and fixed settings are listed in Table~\ref{tab:gine_params}.

\begin{table}[htbp]
\caption{GINE hyperparameter grid and fixed settings.}
\label{tab:gine_params}
\centering
\begin{tabular}{ll}
\toprule
\textbf{Parameter} & \textbf{Value(s)} \\
\midrule
\texttt{hidden\_dim} & \{64, 128\} \\
\texttt{num\_layers} & \{2, 3, 4\} \\
Learning rate & 0.01 (Adam, fixed) \\
LR scheduler & ReduceLROnPlateau (factor 0.5, patience 5) \\
\texttt{max\_epochs} & 200 \\
Early stopping patience & 20 (on val MAE) \\
\texttt{batch\_size} & 64 \\
\texttt{dropout} & 0.5 \\
\bottomrule
\end{tabular}
\end{table}

\end{appendices}

\bibliography{refs}

\end{document}